\documentclass[letterpaper]{article}
\usepackage{natbib,alifeconf}
\usepackage{mdwlist,url}
\usepackage{mathptmx}
\usepackage{moresize}
\usepackage{stmaryrd}
\usepackage{bbm}
\usepackage{amssymb}
\usepackage{graphicx}
\usepackage{verbatim}
\usepackage{subfig}
\usepackage{xcolor,lipsum}
\usepackage{enumitem}

\usepackage{accsupp} 

\makeatletter
\newcommand{\shorteq}{%
  \settowidth{\@tempdima}{-}
  \resizebox{\@tempdima}{\height}{=}%
}

\usepackage{tikz}

\usepackage{calc} 

\usepackage{tikz}

\usepackage[small,compact]{titlesec}

\makeatletter
\newcommand\npar{\@startsection{subsection}{2}{\z@}{-2\p@ \@plus -4\p@ \@minus -4\p@}{-0.5em \@plus -0.22em \@minus -0.1em}{\normalfontnormalsize\bfseries}}
\makeatother

\usepackage{calc} 

\newcommand*\phantomas[3][c]{%
\ifmmode 
\makebox[\widthof{$#2$}][#1]{$#3$}%
\else 
\makebox[\widthof{#2}][#1]{#3}%
\fi 
}

\title{Increased Complexity and Fitness of Artificial Cells that Reproduce Using Spatially Distributed Asynchronous Parallel Processes}

\author{Lance R. Williams$^1$\\
\mbox{}\\
$^1$Department of Computer Science, University of New Mexico, Albuquerque, NM 87131\\ williams@cs.unm.edu}

\begin{document}
\maketitle
\begin{abstract}
Replication time is among the most important components of 
a bacterial cell's reproductive fitness.
Paradoxically, larger cells replicate in less time than smaller cells 
despite the fact that building a larger cell requires increased
quantities of raw materials and energy.
This feat is primarily accomplished by the massive over expression of ribosomes,
which permits translation of mRNA into protein,
the limiting step in reproduction,
to occur at a scale that would be impossible 
were it not for the use of parallel processing.
In computer science, spatial parallelism is the distribution of
work across the nodes of a distributed-memory multicomputer system.
Despite the fact that a non-negligible fraction of artificial life research is grounded
in formulations based on spatially parallel substrates, 
there have been no examples of artificial organisms that use 
spatial parallelism to replicate in less time than smaller organisms.
This paper describes artificial cells defined using a combinator-based
artificial chemistry that replicate in less time than smaller cells.
This is achieved by employing extra copies of programs implementing 
the limiting steps in the process used by the cells to synthesize their component parts.
Significant speedup is demonstrated, despite the increased complexity of 
control and export processes necessitated by the use of a 
parallel replication strategy.
\end{abstract}

\section{Introduction}

On Earth, the origin of life and its evolution into organisms of increasing complexity
was facilitated  by our universe's rich physics, the early Earth's bountiful
solar, geothermal and chemical energy resources,
the massive volume of its prehistoric marine environment, 
and the vast scale of geologic time.
The field of artificial life is premised on the idea that
artificial systems can be constructed which, through simplifications and
other economies of design, are capable of analogous displays of
emergence and open-ended evolution, 
but on vastly shorter time scales within the memories of digital computers.

Any attempt to demonstrate the evolution of organisms of increasing complexity
in an artificial system should begin with an attempt to define complexity itself.
\cite{loyd} gives a list of complexity measures
which he divides into three categories:
1) difficulty of description;
2) difficulty of creation; and
3) degree of organization.
For our purpose, measures based on difficulty of description
suffer from the drawback that random objects have longer descriptions than
non-random objects.
The constraints on complexity imposed by function mean that, 
although complex in the sense that we are attempting to define,
organisms are also far from random.
In Loyd's taxonomy, degree of organization functions as a catch all for measures that attempt to quantify 
the complexity of an object's self-similarity.

\cite{bennett} also reviewed measures of complexity and discussed
their appropriateness for the purpose of quantifying the complexity of living things.
He discussed at length the idea of the multivariate mutual information between different 
parts of an object, which appears in Loyd's list as a measure of degree of organization.
Ultimately, citing the existence of trivial non-living examples,
Bennett discounts self-similarity and proposes a measure, {\it logical depth},
based on difficulty of creation.
Informally, the logical depth of an object is the minimum time required
for any {\it sequential} process to construct the object from some description of the object.
Unfortunately, logical depth can be difficult to estimate in practice, 
since knowing that a process constructs an object from some description in a
given number of steps only establishes an {\it upperbound}.
Perhaps more importantly, the size of the object gives a {\it lowerbound,}
since no sequential process can construct an object composed of
$n$ primitives in less than $n$ steps.
However, any object of size $n$ can be constructed in $n$ steps from a description of length $n$
by a trivial sequential process and objects that can only be constructed this 
way are {\it random}.

Proposed measures of complexity based on difficulty of creation and description
are not specific to self-replicating entities but apply to objects of all kinds.
Moreover, although Bennett's formulation of logical depth is compelling,
there is no implied threshold that, if exceeded, would indicate an
artificial organism's possession of non-trivial complexity.
These considerations suggest two qualities specific to self-replicating 
entities and characteristic of all biological life, yet few (if any) artificial organisms.
The first is that living organisms reproduce by copying and translating
compressed self-descriptions.
The second is that living organisms reproduce in less time than a 
sequential process could construct them from an uncompressed description.
Both are attributes of biological life, but it is the second 
which is the topic of the current paper.

Replication time is among the most important components of 
a bacterial cell's reproductive fitness.
The faster a bacterium replicates, the more likely it is to
outcompete other bacteria for space and nutrients.
It is for this reason that defining complexity as logical depth seems incompatible
with the goal of increasing complexity, since it it seems to imply that 
more complex cells must (by definition) reproduce more slowly.
But bacterial cells are constructed using {\it parallel} not sequential processes.
Using parallel construction processes, it is possible for a bacterial cell
to simultaneously be larger yet require less time to reproduce,
and this is precisely what is observed in nature;
the growth rate of bacterial cells actually increases with size 
 \citep{bremer} and this is only possible because 
 bacterial cells use parallelism at a massive scale to 
 achieve this speedup.
 
\begin{figure}[t]
\begin{center}
\includegraphics[scale = 0.35,bb = 100 50 700 600]{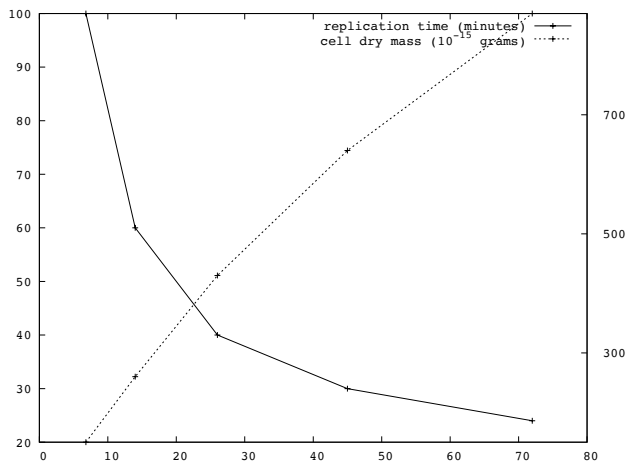}
\caption{{\it E. coli} replication time (left scale is minutes) and dry mass (right scale is $10^{-15}$ grams) 
as functions of number of ribosomes ($\times 10^3$). All numbers are from \cite{milo}.}
\label{ecoli}
\end{center}
\end{figure}

Ribosomes are macromolecular complexes made of rRNA which translate mRNA into protein, 
the most fundamental construction process employed by the living cell.
A bacterium's replication time decreases as the fraction of
its dry mass consisting of rRNA increases. 
In the limit, a single {\it E. coli} can contain $7.2 \times 10^4$ ribosomes
and rRNA can account for fully $37\%$ of its dry mass, achieving a
replication time as short as 24 minutes \citep{milo}.
See Figure \ref{ecoli}.

All living cells have non-trivial algorithmic complexity since they are all many 
orders of magnitude more massive than their genomes,
which function as compressed self-descriptions.
Decompression is accomplished by means of gene expression networks,
which permit genes to be expressed at variable rates and in numbers 
far exceeding their representation in absolute numbers in the genome.
Although ribosomes are made of $rRNA$ not protein,
they are similarly over expressed relative to their representation in the genome.
The {\it E. coli} genome does not have $7.2 \times 10^4$ copies of the gene 
encoding its rRNA;
it only has seven \citep{milo}.

Parallelism is not limited to translation of mRNA into protein by ribosomes.
It is also used to accelerate duplication of the bacterial genome.
This process is implemented by a pair of replication forks that 
copy the circular DNA molecule forming the chromosome of
the mother cell, resulting in a $2 \times$ speedup.
However, \cite{youngren} showed that additional replication forks can
begin construction of the granddaughter cell's chromosomes within 
daughter cells prior to binary fission and before duplication 
of the mother cell's chromosome has completed, 
resulting in additional parallel speedup.

Eukaryotic cells exploit parallelism to an even greater extent,
by incorporation of complex self-replicating components like mitochondria
that reproduce in parallel, and by distribution of their genomes across 
multiple chromosomes, which are copied and exported in parallel.
However, it is indisputable that, with the advent of multicellular organisms,
self-replication by means of parallel processes was taken to an entirely new level.
The existence of organisms as large as elephants and whales
would not be remotely possible were it not for the fact that cell division,
growth and differentiation are inherently parallel 
processes.\footnote{One might consider {\it eusociality} as practiced
by social insects (ants) and some mammals (mole rats) to be a
continuation of this pattern.}

In summary, we believe that the study of self-replication by parallel processes is critical to achieving
the primary goal of the field of artificial life, 
the demonstration of open-ended evolution of artificial organisms of increasing complexity.
Despite this, and despite a long history of research on artificial self-replicating systems \citep{sipper98, freitas04},
there has been relatively little study of systems that replicate
using parallel processes.
The prior work relevant to parallel self-replication can be divided into five categories:
\begin{enumerate} 
\item Artificial organisms that replicate using sequential processes implemented on top of synchronous 
\citep{burks, langton} or asynchronous parallel substrates \citep{nehaniv}.
\item Artificial organisms implemented on top of asynchronous parallel substrates that replicate using 
sequential processes but also make limited use of parallelism \citep{laing, hutton}.
\item Simulations of spatially distributed evolving populations of self-replicating sequential programs \citep{ray, adami1}.
\item Autocatalytic sets defined in artificial chemistries that are not complete organisms because they 
do not  isolate their reactants and products and/or lack the complexity of complete organisms \citep{farmer, nakamura2}.
\item Self-replicating parallel programs in shared-memory multiprocessors \citep{thearling}.
\end{enumerate}

\cite{nakamura2} showed how programs for a universal computer can be encoded 
as a parallel production system hosted on an asynchronous parallel substrate.
He proved important properties of his encoding scheme including guarantees of 
unique solution and freedom from race conditions.
Finally, he described a parallel asynchronous self-replicating system based 
on a self-replicating program encoded using his scheme
which replicates in time logarithmic in program size.
Unfortunately, because no experimental results were described,
this speedup can only be considered a theoretical upperbound and
an implementation on real parallel hardware would be subject to overhead 
like non-zero mixing times of reactants and other factors
not addressed in the paper.

\cite{thearling} described evolution of parallel self-replicating programs 
in a version of Tierra which included a {\it split}
instruction that caused a process to fork into a pair of processes
on a shared-memory multiprocessor.
Unlike spatial parallelism on a distributed-memory multicomputer,
these processes execute in parallel without duplication and distribution 
of the programs defining the processes.
The instruction also modified an index register in a way specific to each 
daughter process so that the address range was divided between them.
The ancestral program was not a {\it quine},
which replicate using a dual process of {\it translation} and {\it copying},
but instead copied itself directly by exploiting its residency 
in the random access shared-memory.
Nor was it sequential;
it contained a single split instruction that reduced execution time by
$2 \times$ relative to a non-parallel copying program.
Although the evolution of programs with reduced execution times was 
demonstrated, it should be noted that 
adding $n$ splits would decrease the execution time
by $2^n \times$ of {\it any} program copying a block of memory of 
size  $2^n$ on this architecture,
and the programs which evolved simply added splits while padding 
with non-operational instructions so that program length 
remained evenly divisible by $2^n$.

\section{Concurrency and Parallelism}

As conceived by \cite{burks}, a self-replicating system consists of four components
\[
A + B + C + D + \phi(A + B+ C + D)
\]
\noindent where $A$ is a {\it translator}, $B$ is a  {\it copier}, $C$ is a {\it controller}, 
$D$ is {\it payload} and $\phi(A + B+ C + D)$ is a {\it description} of $A$, $B$, $C$ and $D$.
Although von Neumann's self-replicating system was formalized as a 
synchronous cellular automaton, the above schema applies equally 
well to an asynchronous system of five {\it actors} which can be viewed 
as an autocatalytic set in an artificial chemistry governed
by two rules:
\begin{eqnarray*}
a: A \!+\! \phi(A \!+\! B\! +\! C\! +\! D) & \!\!\! \!\! \rightarrow \!\!\! \!\!& 2 A \!+ \!\phi(A\! + \! B \!+\!C\! +\! D) \!+\! B \!+ \!C \!+ \!D\\
b: B\! +\! \phi(A\! +\! B\! +\! C\! +\! D) &\!\!\!  \!\! \rightarrow \!\!\! \!\!& B + 2 \phi(A\! +\! B\! +\! C \!+ \!D).
\end{eqnarray*}
These rules have reactants on the left side of the arrow and products on the right.
The first rule translates the description into a set of {\it programs} while
the second rule copies the description.
Each rule describes a {\it subproblem} which must be solved to achieve self-replication.
Subproblems $c$ and $d$ lack reactants or products but might (for example)
recognize when the $a$ and $b$ subproblems have completed, export products
into the daughter cell, and effect fission of an enclosing membrane.
We assume that the times required to solve the $c$ and $d$ subproblems are small  
and (unlike subproblems $a$ and $b$) independent of the 
length of the description, $\phi(A + B + C + D)$.
Nevertheless, $\phi(C)$ and $\phi(D)$ are likely as long as
$\phi(A)$ and $\phi(B)$ and (like them) must be both translated and copied.

In computer science, a {\it concurrent} system consists of subprocesses that can
be executed in different orders, constrained only by {\it data dependencies.}
A data dependency exists when one subprocess provides the input to a second,
in which case the first process must complete before the second can begin.
{\it Concurrency} is a necessary precondition for {\it parallelism}, which is the simultaneous 
execution of subprocesses on different processors.
Parallelism imposes additional constraints on execution order since a resource
being used by one subprocess cannot be simultaneously used by another.
The shared resource in the self-replication problem is the description, 
$\phi(A + B + C + D)$.
It follows that subproblems $a$ and $b$ cannot be solved in parallel, 
which we abbreviate as $a/b$.
Since there are no data dependencies, the execution order of the $a$ and $b$
subproblems is otherwise unconstrained, so that $a \rightarrow b$
and $b \rightarrow a$ are both possible.

The asynchronous system can be reformulated so that parallel solution
of subproblems is possible by splitting the description into four pieces:
\[
A + B + C + D + \phi(A) + \phi(B) + \phi(C) + \phi(D).
\]
The system is still governed by two rules
\begin{eqnarray*}
ax: A + \phi(x) & \rightarrow & A + \phi(x) + x\\
bx: B + \phi(x) & \rightarrow & B + 2 \: \phi(x)
\end{eqnarray*}
where $fx$ is program $f$ and description $x$.
However, there are now eight subproblems, which we refer to as
$aa$, $ab$, $ac$, $ad$, $ba$, $bb$, $bc$ and $bd$.
Because there are no data dependencies between the subproblems, 
they can be solved sequentially in any of $8!$ possible orders, including
\[
aa \rightarrow ab \rightarrow ac \rightarrow ad \rightarrow ba \rightarrow bb \rightarrow bc \rightarrow bd.
\]
To understand the potential for parallel solution of subproblems,
the constraints on simultaneous execution imposed by shared resources
must first be identified.
Execution of a process solving subproblem $f x$ precludes the simultaneous execution 
of processes solving subproblems $f y$ and $g \: x$ since there are
single instances of program $f$ and description $x$.
This imposes ten constraints on simultaneous execution which we 
abbreviate as $fx / fy$ and $ fx / g \: x$.
Nevertheless, even with these constraints, many strategies for parallel solution
of subproblems remain, including
\[
aa  \: | \: ba \rightarrow ab  \: | \: bb \rightarrow ac  \: | \: bc \rightarrow ad \: | \: bd
\]
\noindent which yields a $2 \times$ speedup relative to the sequential strategy.
Distributing the description across multiple actors increased parallelism by 
reducing contention for a shared resource.
However, it also created new opportunities for parallelism which can now be exploited.
This leads us to ask whether or not adding extra copies of $A$, $B$, $\phi(A)$
and $\phi(B)$ might result in further speedup.
Let $n$ be the total number of copies of $A$, $B$, $\phi(A)$ and $\phi(B)$
in an asynchronous self-replicating system:
\[
n (A + \phi(A) + B +\phi(B)) + C + \phi(C) + D + \phi(D).
\]
\noindent The system is governed by two rules:
\begin{eqnarray*}
a_i\:x_j: A_i + \phi(x_j) & \rightarrow & A_i + \phi(x_j) + x_j\\
b_i\:x_j: B_i + \phi(x_j) & \rightarrow & B_i + 2 \: \phi(x_j).
\end{eqnarray*}
It requires the solution of $4 (n + 1)$ subproblems $a_i \: x_j$ and $b_i\: x_j$
to achieve self-replication.
The same constraints on simultaneous execution of the increased number of subproblems apply:
$fx / fy$ and $ fx / gx$.
When $n = 2$, these constraints allow the following strategy
(and many others of equivalent or lesser efficiency)
for parallel solution:
\[
a_0a_0 \: | \: a_1a_1 \: | \: b_0b_0 \: | \: b_1b_1 \rightarrow 
\]\[
a_0b_0 \: | \: a_1b_1 \: | \: b_0a_0 \: | \: b_1a_1 \rightarrow 
a_0c \: | \:  b_0c \: | \: a_1d \: | \: b_1 d.
\]
\noindent Because this strategy contains only three steps,
the speedup is $\frac{4}{3} \times$ relative to the non-redundant parallel 
strategy and $\frac{8}{3} \times$  relative to the sequential strategy.
This system is the first in a series of systems where additional instances of 
the actors which translate and copy can offset the 
replication cost of control and payload actors leading to decreased self-replication time.
Speedup occurs because the work of translating and copying the descriptions 
of the control and payload actors, $\phi(C)$ and $\phi(D)$, 
is divided among all instances of the translate and copy actors, $A$ and $B$.
A hypothetical system containing only translate and copy actors
\[
n (A + \phi(A) + B +\phi(B))
\]
\noindent would not benefit from additional $A$ and $B$ instances, 
since minimum length strategies for all $n$ require two steps:
\[
a_0a_0 \: | \: b_0 b_0  \: | \: \dots \: | \: a_{n-1}a_{n-1} \: | \: b_{n-1} b_{n-1}  \rightarrow 
\]
\[
a_0 b_0 \: | \: b_0 a_0 \: | \: \dots \: | \:
a_{n-1} b_{n-1} \: | \: b_{n-1} a_{n-1}.
\]
Although the example of the asynchronous von Neumann replicator is compelling, 
its lack of data dependencies makes it unusually amenable to speedup via parallelism.
Indeed, it falls into a category of problems often termed, {\it embarrassingly parallel.}
Many problems cannot be decomposed so easily.
Although data dependencies do not necessarily preclude parallel speedup,
they sometimes allow only a more limited form of parallelism termed {\it pipeline parallelism.}
Consider an asynchronous self-replicating system governed by the following two rules:
\begin{eqnarray*}
a_ix_j: A_i + \phi(x_j) & \rightarrow & A_i + 2 \: \tilde{\phi}(x_j)\\
b_ix_j: B_i + 2 \: \tilde{\phi}(x_j) & \rightarrow & B_i + x_j + 2 \: \phi(x_j).
\end{eqnarray*}
The first rule consumes a description $\phi(x_j)$ and produces a pair of 
{\it reversed} descriptions $2 \: \tilde{\phi}(x_j)$.
The second rule consumes a pair of reversed descriptions and produces 
a program $x_j$ and two descriptions $2 \: \phi(x_j)$.
In effect, the second rule simultaneously translates and copies.
Although this process might seem contrived,
it is very natural when programs and descriptions are represented 
as stacks that can be copied in a two stage
process using a pair of stacks as an intermediate representation.
More importantly, it is the process used by the artificial cell 
described later in the paper.

A self-replicating system using this process for translation and copying
contains four data dependencies not present in the embarrassingly parallel system.
These can be abbreviated as $bx > ax$, since for all $x$ subproblem $bx$ 
cannot be solved before subproblem $ax$ has completed.
Despite these constraints, for $n=2$, the following strategy is possible
\[
a_0a_0 \: | \: a_1a_1 \rightarrow a_0b_0 \: | \: a_1b_1 \: | \: b_0a_0 \: | \: b_1 a_1 \rightarrow 
\]\[
b_0 b_0 \: | \: b_1 b_1 \: | \: a_0c \: | \: a_1 d \rightarrow b_0c \: | \: b_1d.
\]
\noindent This strategy yields a $\frac{5}{2} \times$ speedup relative to the sequential strategy.
Although not quite as large as the $\frac{8}{3} \times$ speedup observed in the
embarrassingly parallel system, the speedup in the parallel pipelined system
is still significant.

\section{Artificial Chemistry of Program Fragments}

An artificial chemistry is a system of constructible objects \citep{fontana99}.
Elemental objects called {\it atoms} are combined to form
more complex objects by reaction rules.
When objects are embedded in a two
dimensional space, and are subject to diffusion,
the artificial chemistry is a {\it bespoke physics} \citep{bespoke}.
A bespoke physics serves as an abstract interface to a substrate comprised of 
asynchronous cellular automata \citep{priese} which can be simulated 
in parallel on a distributed-memory multicomputer.
Realism can be increased by assuming that
atoms cannot be created or destroyed, 
so that mass is conserved \citep{pietro}.
In a typical artificial chemistry, objects are symbols, strings or graphs 
and the rules which construct them are immutable.
However, when objects can represent {\it programs} and the immutable rules
implement an {\it interpreter}, an artificial chemistry can (in effect) construct
and apply new rules \citep{fontana99, tomita07, hickinbotham, marius14}.
These artificial chemistries have increased expressive power.

In functional programming, {\it monads} are an abstract data type
representing {\it program fragments} \citep{moggi}.
The monad interface allows program fragments to be composed and applied to values.
By making control idioms implicit in data types, 
monads make it possible to build simple programs exhibiting complex behaviors,
{\it e.g.,} backtracking.
This makes them a powerful tool for defining highly expressive artificial chemistries.
In prior work \cite{pop} described an artificial chemistry where atoms are 
monadic combinators and programs implementing non-deterministic 
rules are constructed objects.
The atoms, which cannot be created or destroyed,
are embedded in a two dimensional space and subject to diffusion.
Programs composed of combinators can construct both spatially distributed objects,
and non-distributed objects of increased mass.
Programs themselves are objects of this second type.
Objects of increased mass diffuse more slowly,
reflecting the real cost of data transport in the asynchronous 
parallel substrate.
Upon completion, programs report the number of operations they executed.
Because this number is a sum over all paths of execution, the real cost of
simulating non-deterministic rules on the substrate is paid for.
The asynchronous updates of local state needed to simulate both the execution 
of spatially distributed processes running for different amounts of time and 
the diffusion of objects of unequal mass are uniformly managed
using the Gillespie algorithm \citep{gillespie}.

\section{Parallel Export}
{\it Artificial organisms} are more complex than self-replicating systems since, 
in addition to making their own parts, they must also segregate 
their parts from the parts of other organisms \citep{mcmullin}.
Given a device which implements this segregation, 
the system can become a full-fledged organism by adding one or more instances of a program $E$
that moves the products of the mother's translation and copying processes
from mother to daughter:
\[
n (A \!+\! \phi(A) \! +\! B  \!+ \! \phi(B)) \! + \! C \! + \! \phi(C) \! + \! D \! + \! \phi(D) \! + \! m (E \! + \! \phi(E)).
\]
The organism requires an additional rule governing export:
\begin{eqnarray*}
e_i\:x_j: E_i + x_j +  \phi(x_j)  & \rightarrow & E_i + x_j^\prime + \phi(x_j)^\prime.
\end{eqnarray*}
\noindent where $x_j^\prime$ and $\phi(x_j)^\prime$ are a program
and its description after export to the daughter organism.
Because export of products can only happen after products are completed,
two additional constraints on execution order are needed,
$e \: x > a \: x$  and $e \: x > b \: x$.
Since $E$ exports both a program and its description, the number of steps 
required to solve subproblem $e \: x$ is twice the number
required to solve $a \: x$ or $b \: x$.
However, these steps are of unequal size,
and although the export process is sequential when $m=1$,
the value of $m$ needed to avoid an export bottleneck 
depends on the relative cost of primitive synthesis 
and export operations.

Segregation of an organism's component parts from those of other organisms 
can be accomplished in different ways.
If an organism is defined solely as a pattern of states inside a compact,
connected region of a parallel substrate, then distance alone can serve this purpose.
A daughter organism can be constructed at some offset
relative to its mother using the device of a construction arm \citep{burks, langton, nehaniv}.
Unfortunately, the use of this device eliminates the possibility of
parallel speedup since it acts as a sequential bottleneck.
Irrespective of the fact that it is spatially distributed and uses
pipeline parallelism in its transmission of signals,
the von Neumann automaton, considered as a whole,
is sequential since the daughter automaton is constructed one site at a time
at the tip of the moving construction arm.

Moveable organisms require a different solution.
The most straightforward is to represent them as the connected components 
of a graph embedded in the substrate.
The work of \cite{hutton} serves as an excellent example.
Although this groundbreaking work has influenced 
the thinking of this author and many others,
Hutton's ``cells'' are probably more accurately described as 
``molecules'' since they consist of a few dozen atoms linked by bonds, 
lack membranes and genomes, and are copied by the reaction 
rules of a complex artificial chemistry designed solely for this purpose.
The reaction rules implement a sequential and 
deterministic process that results in two half-sized daughters,
which are then completed by means of a growth process 
that is concurrent and spatially parallel.

In living cells, segregation is accomplished using membranes.
Viewed abstractly, membranes are just topological spheres that partition space 
into two disjoint volumes called {\it inside} and {\it outside}.
The computational problems associated with the use of membranes
as a segregation device were discussed at length in \cite{newcastle}.
Although space considerations preclude a detailed reprise of that discussion here,
its conclusion was that using membranes to partition space is 
deceptively complicated, yet also completely unnecessary since simpler methods
for representing membership in compact, spatially embedded sets 
of objects exist.

\cite{newcastle} describes one such method in detail.
The footprint of a {\it roving pile} is a connected component of a two dimensional lattice graph.
In the combinator-based artificial chemistry described in \cite{pop},
actors can form groups which diffuse as a single unit.
Groups in the footprint serve as bases for stacks of groups
which form the contents of the pile.
Base groups with one or more missing edges in the lattice graph
form the footprint's {\it boundary.}
Programs in the same stack execute concurrently but not in parallel;
they compete for a shared processor resource in zero sum fashion.
However, programs in different stacks in the same pile execute in parallel.
So that piles can move and grow, and so that objects within piles 
can freely mix, groups in piles undergo diffusion.
The footprint's shape evolves because groups on the boundary 
can move inwards or outwards subject only to the constraint that such movement
cannot disconnect the pile's footprint.
Although implemented as connected components of a lattice graph,
the footprints of a roving pile can be visualized as 
planar immersions of two-dimensional surfaces with boundaries,
and they are depicted this way in the figures of this paper.
Observation of a working implementation shows that roving piles 
remain flat (low average stack height) and connected.
Smaller piles constantly evolve in shape while rapidly moving around the 
space on random walks.
Larger piles extend and retract pseudopod-like extensions 
but are less mobile in aggregate.
In summary, roving piles look a lot like living cells.

\begin{figure}[t]
\begin{center}
\includegraphics[scale = 0.3,bb = 100 0 700 150]{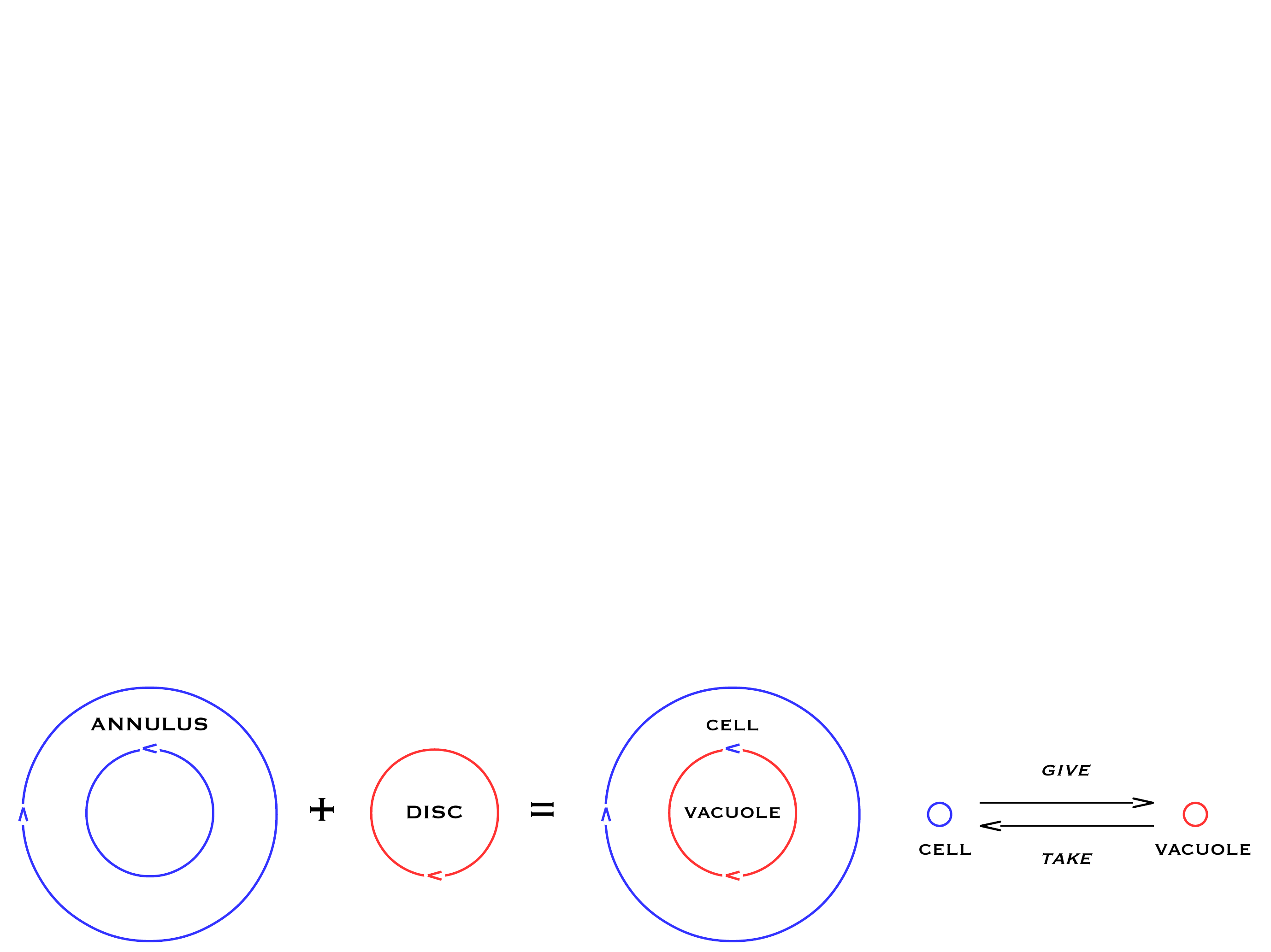}
\caption{Topologically identifying the inner boundary of an {\it annulus} and 
the boundary of a {\it disc} creates the footprint of a roving pile
with two compartments called {\it cell} and {\it vacuole}.
The boundary between compartments is {\it selectively permeable.}
Transport is mediated by a pair of combinators that act as a 
{\it transport interface} (right).
These combinators modify {\it group state.}
Groups in the {\it give} state can move (by diffusion) from cell to vacuole
while groups in the {\it take} state can move in
the opposite direction.}
\label{vacuole}
\end{center}
\end{figure}

\begin{figure}[t]
\begin{center}
\includegraphics[scale = 0.3,bb = 100 0 700 300]{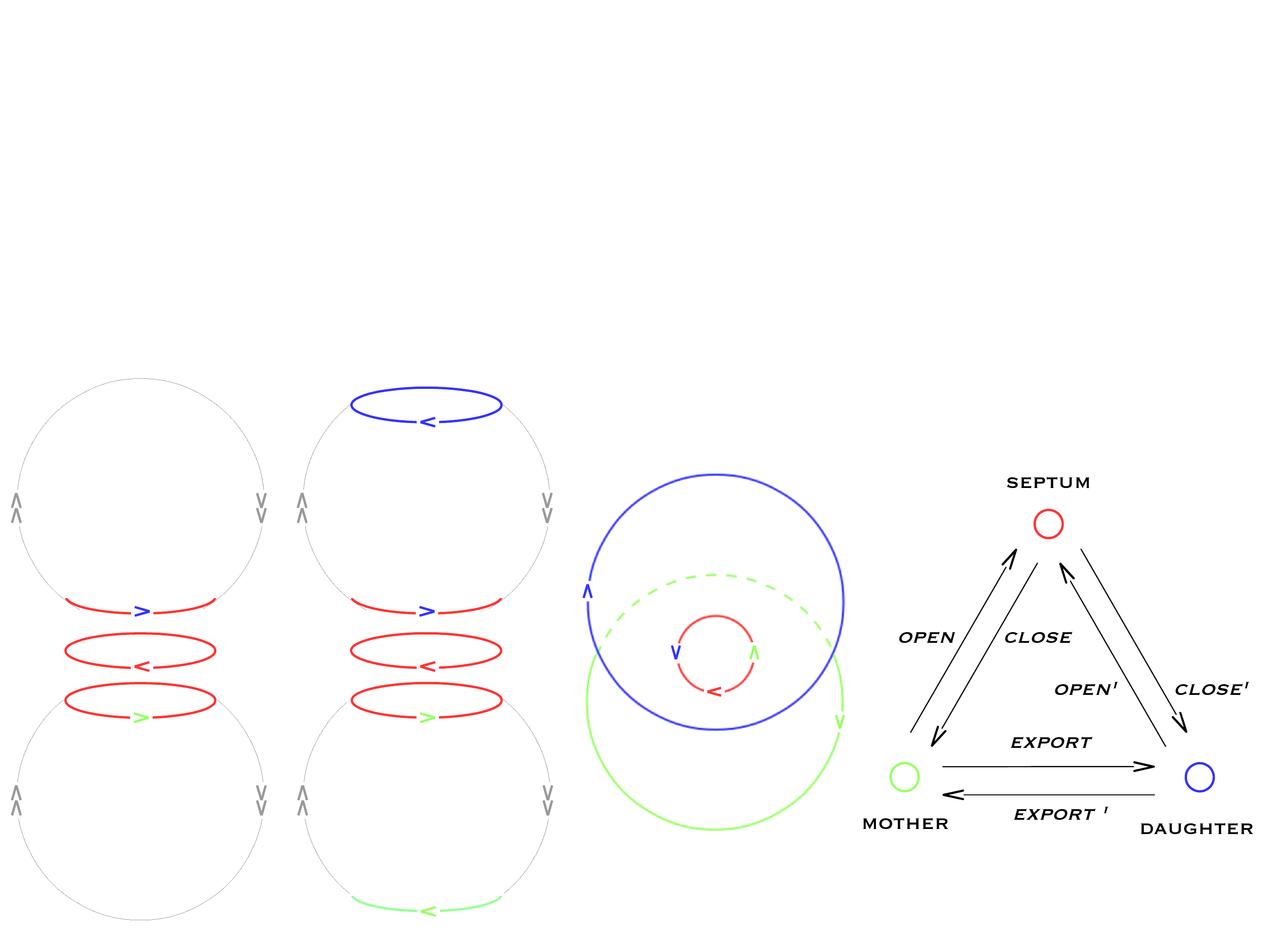}
\caption{Exploded view in three dimensions of a cell in the process of binary fission showing
a pair of circular apertures that pierce daughter cell membranes and 
the {\it septum} that divides the cell volume (left).
The pair of apertures and the boundary of the septum are topologically identified
along the course of the {\it contractile ring} (red).
Adding a second circular aperture to each membrane at the end opposite
the septum (blue and green) creates a {\it topological complex} that can be flattened,  
{\it i.e., immersed} in the plane.
This complex is the footprint of a roving pile with three compartments
called  {\it mother}, {\it daughter} and {\it septum}.
The boundary shared by the three compartments (red)
is selectively permeable.
Export from mother to daughter is mediated by a set of combinators
that act as an {\it export interface} (right).
{\it Open} (or {\it close}) change group states to {\it open} (or {\it close}),
conferring permission to diffuse from mother to septum (or {\it vice versa}).
This causes the area of the septum to increase (or decrease),
which increases (or decreases) the length of the shared boundary,
controlling the rate of diffusion from mother to daughter
of groups in the {\it export} state.}
\label{interface}
\end{center}
\end{figure}

\begin{figure}[t]
\begin{center}
\includegraphics[scale = 0.3,bb = 100 0 700 300]{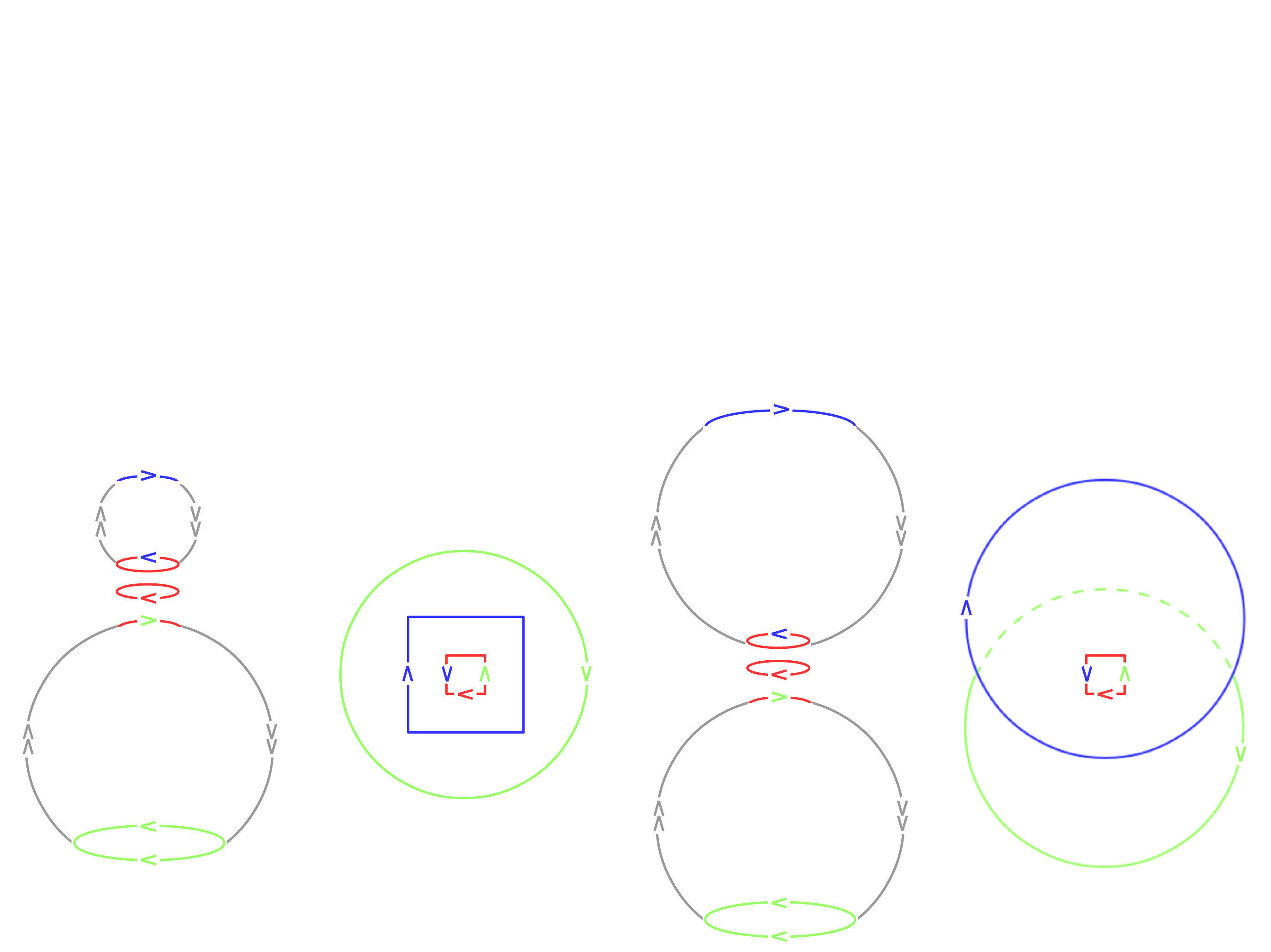}
\caption{Exploded view of mother cell with {\it bud} and planar immersion of complex (left).
The septum's footprint is a square of size $1 \times 1$, 
while the mother and daughter cell's footprints are annuli with
square inner boundaries of size $1 \times 1$ (red).
The outer boundary of the daughter cell's footprint is a square of
size $3 \times 3$ (blue).
Significantly, the footprints of the daughter and septum can be created
using $O(1)$ operations in a single Moore neighborhood 
of the mother's footprint.
Exploded view of mother and daughter cell immediately prior to fission
and planar immersion of complex (right).
The footprint of the septum is a square of size $1 \times 1$, 
while the footprints of the mother and daughter are annuli
with square inner boundaries of size $1 \times 1$.
The septum can be ejected and the holes in mother and
daughter filled using $O(1)$ operations in a 
single Moore neighborhood.}
\label{budding}
\end{center}
\end{figure}

To achieve the results reported in the present paper,
the functionality of roving piles was significantly enhanced beyond
that described in \cite{newcastle}.
The artificial cells described in this paper are hosted in piles with 
multiple compartments separated by selectively permeable boundaries;
see Figure \ref{vacuole}.
Using this device, it is possible to implement a parallel export process that
resembles binary fission in living cells.
See Figure \ref{interface} and Figure \ref{budding}.


\begin{figure}[t]
\begin{center}
\includegraphics[scale = 0.55,bb = 175 10 700 350]{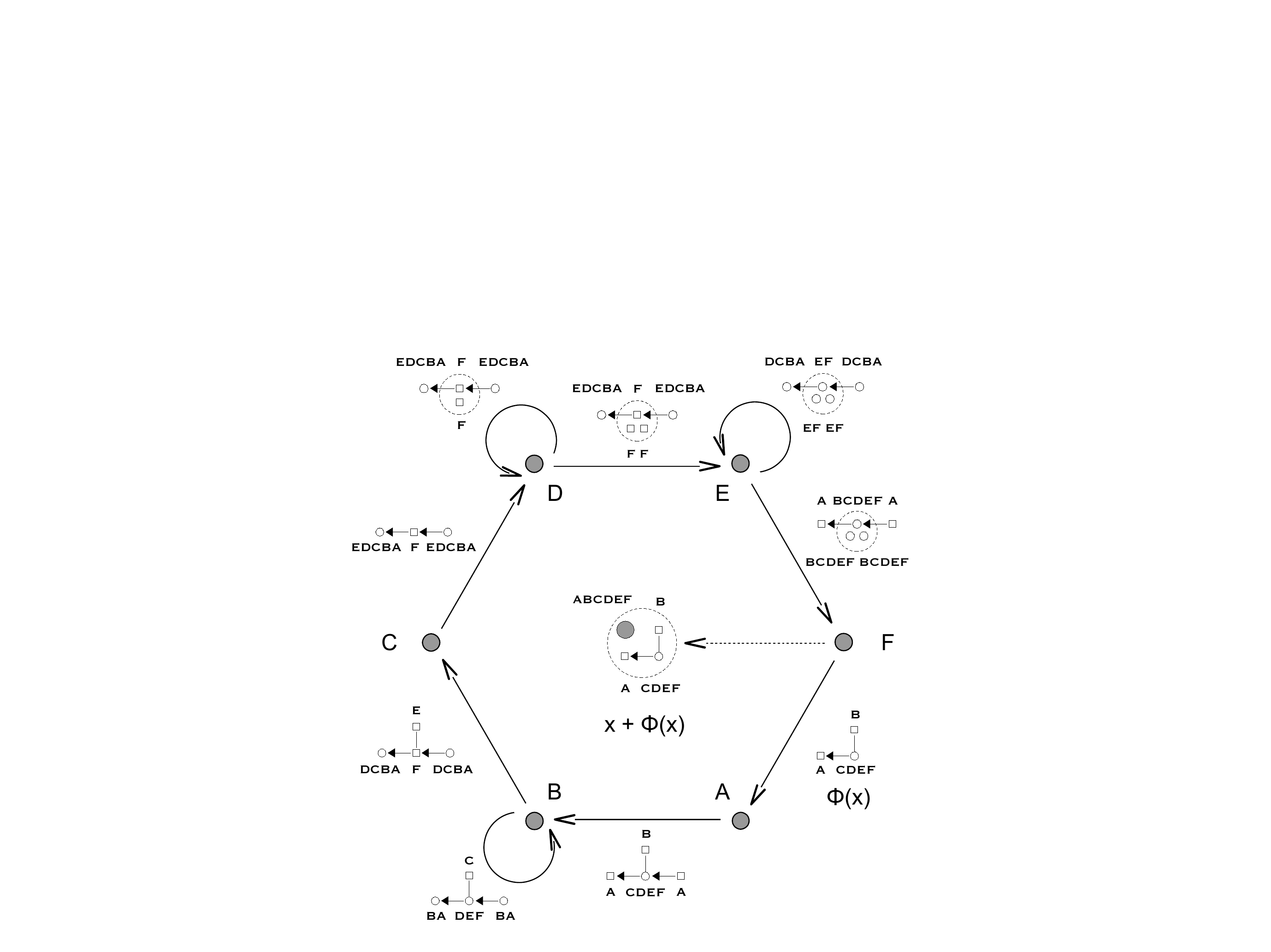}
\caption{Six stage parallel pipeline for translating and copying
zipper representation of description $\phi(x)$ showing changes to zipper conformation 
effected by programs $A$-$F$.
Unlike the pipeline described in \cite{newcastle},
this pipeline produces 
both program $x$ and description $\phi(x)$ in a single cycle.
This permits export to be implemented as a parallel process.}
\label{lucifer}
\end{center}
\end{figure}

\begin{figure}[t]
\begin{center}
\includegraphics[scale = 0.35,bb = 75 0 700 500]{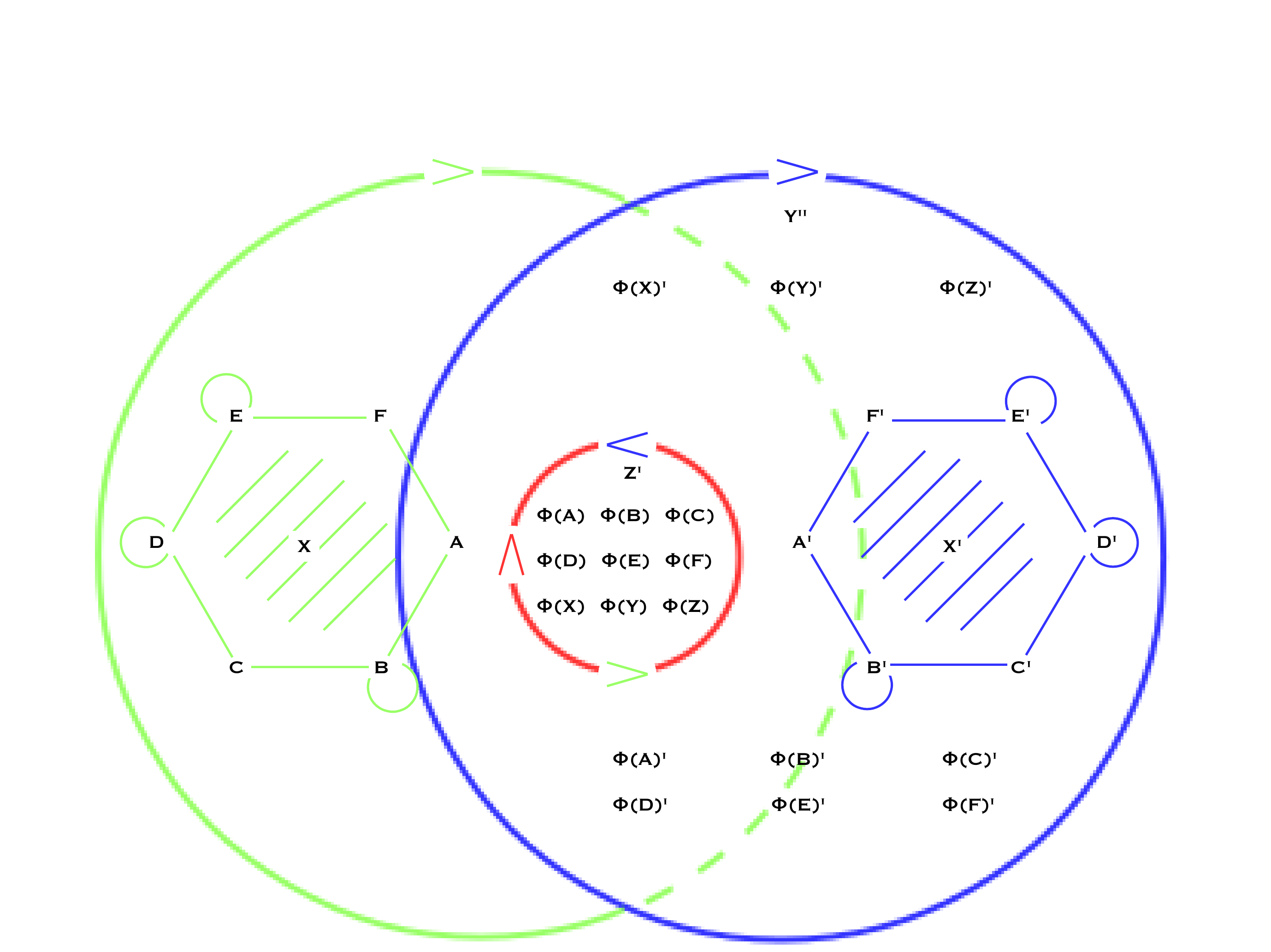}
\caption{Schematic diagram of artificial cell.
Mother (green) and daughter (blue) are depicted immediately prior to fission.
After duplication by pipeline implemented by methods $A$-$F$,
zippers $\phi(x)$ representing the genome migrate to the
septum (red) to prevent further duplication.
Synthesis of $Y''$ by $A'$-$F'$ triggers $Z^{\:\prime}$, 
which returns zippers to the mother while closing the septum (red).
Parallel speedup results from extra copies of $B$ and $E$.}
\label{fig:satori}
\end{center}
\end{figure}

\section{Parallel Speedup in Artificial Cells}

As \cite{mcmullin} noted, an artificial cell must accomplish two tasks.
First, it must reproduce its own components.
Second, the components and their duplicates must be spatially organized into separate cells.
If the organization process is symmetrical, then the separate cells are both {\it daughters.}
However, if the organization process is asymmetrical (as in yeast which reproduce by budding) 
then the two cells are distinguishable and can be called {\it mother} and {\it daughter.}
We implemented an artificial cell that accomplishes the first
of these tasks using a parallel pipeline consisting of six programs $A$-$F$
that translate and copy program descriptions;
see Figure \ref{lucifer}.
The second task is accomplished by programs $X$, $Y$ and $Z$,
which manage the organization of components and 
duplicates into mother and daughter cells.
$X$ consists of a short executable segment followed by a much longer non-executable segment.
When it can do so without splitting the mother's pile, $X$ creates a {\it bud} and
moves itself to the newly created daughter's footprint; see Figure \ref{budding} (left).
It then {\it smashes} itself within the daughter's footprint,
creating the daughter's {\it cytosol} from the primitives comprising its non-executable segment.
The cytosol contains the minimum set of primitive combinators necessary for self-replication.
These are consumed by the $A$-$F$ pipeline during synthesis and replenished by imports to the pile.
$Z$ closes the septum after verifying the daughter cell's viability
and $Y$ is a short non-executable program that is manufactured by the daughter 
just to prove its viability to $Z$.
The lengths of the nine programs comprising the artificial cell
are given in Table \ref{table:balance}.
\begin{table}[ht]
\caption{Program lengths (combinators).}
\centering
\begin{tabular}{| c | c | c | c  | c | c | c | c | c |}
\hline
A  & B & C & D  & E  &  F  & X & Y & Z\\
\hline
42 & 62 & 55 & 63 & 74  &  92  & 89 & 8 & 43\\
\hline
\end{tabular}
\label{table:balance}
\end{table}
A {\it zipper} is an implementation of a data structure that can be traversed
and modified without mutation \citep{zipper}.
All zippers consist of three parts.
The {\it front} and {\it back} represent the parts of the data structure 
that either 1) have already been traversed; or 2) have yet to be traversed.
The {\it focus} is an item between the front and back that can be inspected or replaced.
The input to the synthesis pipeline is a zipper representation of $\phi(x)$.
A pair of reversed program descriptions $2\: \tilde{\phi}(x)$
can be constructed by traversing this zipper.
The process begins when $A$ copies the front of the input zipper.
The front initially consists of a single combinator.
In each step of the traversal, $B$ pushes the focus onto 
the mother front and a primitive from the neighborhood
(with matching type) onto the daughter front;
the back is then popped to create a new focus.
This is repeated until the back consists of a single combinator
and the mother and daughter fronts each hold a reversed description $\tilde{\phi}(x)$.
These are (in turn) reversed by a second traversal of the zipper in 
the opposite direction by $E$ yielding $\phi(x)$ and $\phi(x)^\prime$.
Significantly, during this traversal, $E$ also makes an instance of $x^\prime$.
The time required for synthesis by the $A$-$F$ pipeline is dominated by 
stages $B$ and $E$ which require time proportional to the
length of $\phi(x)$.
All other stages require constant time.

The last stage in the synthesis pipeline is implemented by $F$, 
which constructs a group in the {\it export} state containing $x^\prime$ and 
$\phi(x)^\prime$ and restores the mother zipper $\phi(x)$ 
to the conformation needed by $A$,
which implements the first stage in the pipeline.
However, $A$ will not actually see it until after replication completes,
since $F$ also changes the state of the group containing the
mother zipper $\phi(x)$ to {\it open}.
This causes it to migrate to the septum which improves the efficiency of
the synthesis process by reducing contention
while also preventing 
the synthesis of unneeded instances of $x^\prime$ and $\phi(x)^\prime$.
Translation of $\phi(Y)^\prime$ by $A'$-$F'$ in the daughter cell 
causes $Z^{\:\prime}$ to migrate to the septum.
Once in the septum, $Z^{\:\prime}$ returns the zippers $\phi(x)$ stashed
there to the mother, where they reenter the $A$-$F$ pipeline, 
restarting the mother's self-replication process.
Of course, by this point, the daughter's self-replication process is already in progress;
see Figure \ref{fig:satori}.
When $Z^{\:\prime}$ is the last actor remaining in the septum,
it closes the septum and ejects itself from the pile, 
completing the final operation in the 
construction of the daughter cell;
see Figure \ref{budding} (right). 

It should be noted that this differs from the process described in \cite{newcastle} 
because the $E$ stage of the pipeline in that process reversed but did not copy.
The advantage of the new pipeline is that both $x^\prime$ and $\phi(x)^\prime$
are produced in a single cycle.
Significantly, it is this modification which permits export to be implemented as a parallel process.
To appreciate this, recall that the cell described in \cite{newcastle} used a 
set of actors contained in a single group as a checklist to ensure that the 
daughter received the full complement of programs and descriptions.
The fact that the checklist was non-distributed yet needed to be maintained 
by any programs implementing the export process necessitated that 
the process be sequential.
The fact that $x^\prime$ and $\phi(x)^\prime$ are exported in a single group 
in the new cell means that the daughter can 
demonstrate that it has received the full complement of 
programs and descriptions simply by
using its synthesis pipeline to translate and copy the description 
of a short dummy program $\phi(Y)^\prime$.
Because programs and descriptions are exported in pairs,
its ability to do this demonstrates that it possesses not only 
$A^\prime$-$F^\prime$, $X^\prime$ and $\phi(Y)^\prime$,
but also $\phi(A)^\prime$-$\phi(F)^\prime$, $\phi(X)^\prime$ and $Y^\prime$.
Finally, because $Z{\:^\prime}$ closes the septum but migrates there from the daughter, 
fission cannot occur unless the daughter also possesses $\phi(Z)^\prime$.

\begin{figure}[t]
\begin{center}
\includegraphics[scale = 0.35,bb = 100 50 700 550]{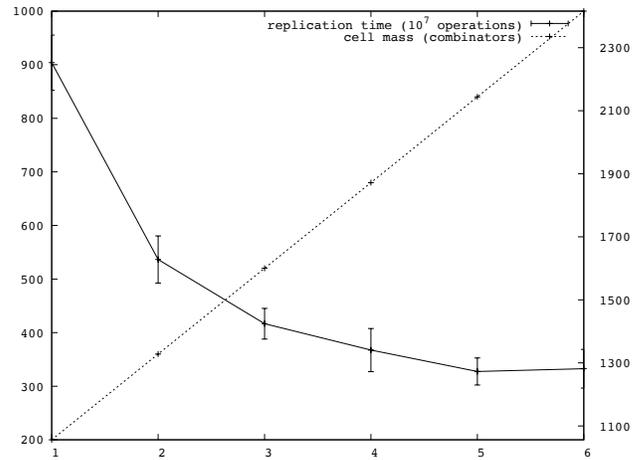}
\caption{Artificial cell replication time (left scale is $10^7$ operations) and mass (right scale is combinators) as 
functions of number of copies of $B$ and $E$.
Error bars show $\pm \sigma$.}
\label{experiment}
\end{center}
\end{figure}

The artificial cell's replication time as a function of the number of copies $(n)$ of
the $B$ and $E$ programs and descriptions was measured experimentally.
All trials were repeated ten times.
Significant decrease in replication time relative to $n = 1$ was observed for 
$1 < n < 7$, consistent with parallel speedup.
This speedup occurred despite the increased size of the artificial cells; 
see Figure \ref{experiment}.
Replication time continued to decrease with increasing $n$ before reaching 
a minimum value at $n = 5$ followed by a slight increase at $n=6$.

To ascertain the speedup relative to a sequential implementation of an otherwise identical process,
an artificial cell was constructed that combined the $A$-$F$ programs
into a single method together with programs for construction of the 
daughter cell's cytosol and export.
This method and its description formed a single group.
Because of its simplified export process,
this sequential artificial cell had a significantly reduced size (734 combinators).
Despite its reduced size,
the sequential cell's replication time ($405 \times 10^7$ operations)
exceeded the replication times of much larger
parallel cells ($> 1872$ combinators) for $n > 3$.
This demonstrates that spatial parallelism can pay for the increased complexity
and runtime overhead associated with its use,
resulting in artificial cells of increased complexity and fitness.

The export of redundant genes from mother to daughter is not completely reliable.
Unlike \cite{nakamura2}, the parallel computation described in this 
paper is subject to race conditions that can potentially affect its result.
Because they are unnecessary for synthesis of $Y''$,
it sometimes happens that unfinished copies of extra $B$ and $E$ 
are in the mother when the septum closes.
Since $X$ is significantly longer than both $B$ and $E$, this happens only rarely.
However, when it does, the number of copies of the affected gene in the 
current daughter decreases
by one and the number in the next daughter (its younger sister) increases by one.
It follows that, in a large population, there will be some variation in $n_B$ and $n_E$,
the number of copies of the $B$ and $E$ genes.
Given sufficient time, natural selection would presumably cause the means
to converge to values maximizing the fitness tradeoff between
increased size (bad) and increased parallelism (good).
Characterizing the distribution of $n_B$ and $n_E$ in a large population,
and potentially observing their convergence over time to optimum values, 
is a subject for future work.

\section{Conclusion}

Reproduction by spatially distributed asynchronous parallel processes is essential to the 
long term goal
of demonstrating open-ended evolution of artificial organisms of increasing complexity.
Despite this fact and the extensive literature on self-replicating systems,
there has been little research on systems that replicate
using parallel processes.
This paper described artificial cells that replicate using spatially distributed
asynchronous parallel processes defined using a combinator-based
artificial chemistry.
Larger cells that reproduce in less time than smaller cells were demonstrated.
This was achieved by adding extra copies of programs implementing the
limiting steps in the parallel pipelined process used by the cells to synthesize 
their component parts.
Significant speedup was observed, despite the increased complexity of 
control and export processes necessitated by the use of a 
parallel replication strategy.
The similarity between the parallel speedup observed in artificial cells
with increased numbers of programs implementing limiting
steps in their component synthesis pipelines and that of bacterial cells 
with increased numbers of ribosomes is noteworthy.

{\small
\bibliographystyle{apalike}
\bibliography{alife21} 
}

\end{document}